\documentclass{article}

\PassOptionsToPackage{numbers, compress}{natbib}

\usepackage[preprint]{neurips_2022}

\usepackage[utf8]{inputenc} 
\usepackage[T1]{fontenc}    
\usepackage{hyperref}       
\usepackage{url}            
\usepackage{booktabs}       
\usepackage{amsfonts}       
\usepackage{nicefrac}       
\usepackage{microtype}      
\usepackage{xcolor}         

\usepackage{amsmath,amssymb,amsfonts}
\usepackage{algorithmic}
\usepackage{algorithm}
\usepackage{graphicx}
\usepackage{textcomp}
\usepackage{xcolor}
\usepackage{caption}
\usepackage{subcaption}
\usepackage{url}
\usepackage{hyperref}

\usepackage{bbm}

\title{FASTER-CE: Fast, Sparse, Transparent, and Robust Counterfactual Explanations}

\author{%
  Shubham Sharma \\
  University of Texas at Austin\\
  \texttt{shubham\_sharma@utexas.edu} \\
  \And
  Alan H. Gee\\
  Amira Learning\\
  \texttt{alan.gee@amiralearning.com}\\
  \And
  Jette Henderson \\
  CognitiveScale\\
  \texttt{jhenderson@cognitivescale.com} \\
  \And
  Joydeep Ghosh \\
  University of Texas at Austin\\
  \texttt{jghosh@utexas.edu} \\
}

\begin{document}

\maketitle

\begin{abstract}
  Counterfactual explanations have substantially increased in popularity in the past few years as a useful human-centric way of understanding individual black-box model predictions. While several properties desired of high-quality counterfactuals have been identified in the literature, three crucial concerns: the speed of explanation generation, robustness/sensitivity and succinctness of explanations (sparsity)  have been relatively unexplored. In this paper, we present FASTER-CE: a novel set of algorithms to generate fast, sparse, and robust counterfactual explanations. The key idea is to efficiently find promising search directions for counterfactuals in a latent space that is specified via an autoencoder. These directions are determined based on gradients with respect to each of the original input features as well as of the target, as estimated in the latent space. The ability to 
quickly examine combinations of the most promising gradient directions as well as to incorporate additional user-defined constraints allows us to generate multiple counterfactual explanations that are sparse, realistic, and robust to input manipulations. Through experiments on three datasets of varied complexities, we show that FASTER-CE is not only much faster than other state of the art methods for generating multiple explanations but also is significantly superior when considering a larger set of desirable (and often conflicting) properties. Specifically we present results across multiple performance metrics: sparsity, proximity, validity, speed of generation, and the robustness of explanations, to highlight the capabilities of the FASTER-CE family.
\end{abstract}

\section{Introduction}

  
\begin{figure*}
  \includegraphics[width=\textwidth]{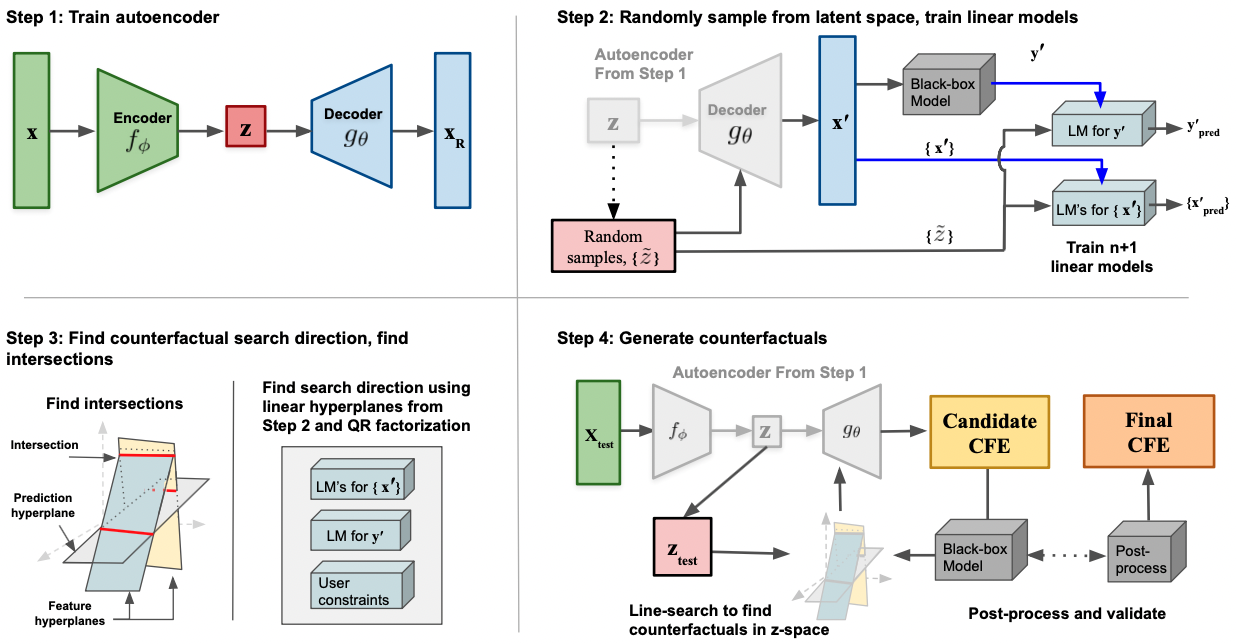}
  \caption{The FASTER-CE method: An autoencoder is trained on the input dataset. Then, the latent space is extensively sampled to generate random latent space vectors that correspond to random decoder outputs, and for every random perturbation, the value of every feature from the decoder output, and the black-box prediction for the decoder output are noted. For every feature in the dataset and for the black-box prediction, a linear model is trained with the latent space vectors as inputs to the linear model and the feature or the black-box prediction as the output. After finding these linear hyperplanes associated with each feature and the black-box prediction, user-defined constraints are incorporated and a combination of QR factorization with respect to the normal vectors of the hyperplanes and finding interesections between hyperplanes gives us a direction to move to find candidate explanations. Multiple candidates are found for the test point for which we need an explanation and checked for validity and post-process, and a set of final explanations is returned.}
  \label{fig:FASTERCE}
\end{figure*}

As machine learning models get deployed in high-stake applications such as finance and healthcare, providing human-centric explanations for the decisions made by such models is becoming increasingly important. Not surprisingly, a wide variety of explanation types with associated algorithms have been proposed and deployed recently \citep{towards}, including feature attribution methods, prototypes and counterfactuals. Among these, counterfactual explanations have found increased traction as a useful, human-centric way of  providing individual-level explanations for the decision of a black-box model. Originally introduced in \citep{wachter2017counterfactual}, counterfactual explanations attempt to answer the question: "What would have to change in the input to change the outcome of the model?".
Such an explanation can be used as a means of actionable recourse for an end-user subjected to a model decision. The original definition \citep{wachter2017counterfactual} focussed on finding a valid counterfactual as close to the  input as possible, where "valid" means that the class label given by the model is different for the counterfactual.
However it was soon apparent that useful counterfactuals  exhibit several other desirable characteristics as well \citep{verma2020counterfactual,mothilal2020explaining,sharma2020certifai}. 
The sparsity of an explanation, i.e., the number of attributes/features that change
to obtain the counterfactual, is one such property. For an end-user, changing just one attribute may be easier and more understandable compared to changing multiple features in order to receive a different outcome. However, including a sparsity constraint in existing methods that provide black-box counterfactual explanations as an objective is fairly non-trivial, since many of these methods rely on gradient descent \citep{wachter2017counterfactual,mothilal2020explaining,balasubramanian2020latent} and an  L0 penalty is non-differentiable, so surrogates may be needed.  Another key concern is the robustness of counterfactual explanations. Recently, it has been shown that counterfactual explanations  are vulnerable to input perturbations as well as adversarial attacks \citep{slack2021counterfactual}.  The realism and actionability of an explanation is also desirable.  Counterfactuals that are   "in-distribution" are preferred, so are those where the indicated change is
feasible. For example, a counterfactual that states that one needs to decrease her age to get a more preferred outcome, indicates an infeasible change and hence is not an actionable means of recourse for that end-user.

However, all these desirable properties can be at odds with each other. For example, the nearest point with a different outcome may be in an infeasible change direction.  Moreover, given multiple counterfactuals for a given input, the most actionable choice for recourse can be subjective and thus should be an end-user's choice. We have also come across  applications such as automated approval for medical procedures, where many decisions are being made each hour and explanations are needed for each one. In such cases, the speed of explanation generation becomes crucial because of the (near) real-time business requirements, even though it may cause one to miss some counterfactuals that could have been more actionable.

In this paper, we propose FASTER-CE: FAst, Sparse, TransparEnt, and Robust Counterfactual Explanations. The FASTER-CE method is shown in Figure \ref{fig:FASTERCE}. First, an autoencoder is trained on the input dataset. Then, the latent space of this autoencoder is sampled from to generate a set of latent vectors $\{\tilde{\textbf{z}}\}$ and corresponding decoder outputs $\{\textbf{x'}\}$. Then, we train a set of linear models mapping the latent vectors
to  each individual input feature now considered as the target, and one more linear models that uses the black-box predictions $y'$ of the decoder output as target. Furthermore, using QR factorization to incorporate user-defined constraints and by finding the intersection of hyperplanes defined by the linear models, we identify promising direction(s) of counterfactual search and use line search to generate a set of candidate counterfactual explanations for a given test data point $\textbf{x}_{test}$. The validity of these explanations is checked, post-processing is carried out if need be to induce sparsity, and the set of valid counterfactual explanations is returned.

There are multiple advantages of using this process to generate counterfactual explanations. A reasonably accurate autoencoder provides a lower dimensional representation of the data in which the search for counterfactuals is quicker, more realistic and robust. Moreover, training the autoencoder and the linear models is a one-time process. Finding counterfactual explanations for any future input then is simply based on a projection of its latent representation onto the intersection of hyperplanes and moving along specific directions. This is significantly faster than performing gradient descent \citep{mothilal2020explaining}, using an evolutionary algorithm \citep{sharma2020certifai}, or the many existing methods that require an optimization procedure for every test point. Through QR factorization, we are able to generate the sparsest explanations, i.e., where only one feature changes (when feasible). We  can also generate multiple explanations efficiently with multiple features changing, if required. Finally, the use of linear models makes it easier
to specify a margin around the boundary in which an explanation should not exist, thereby allowing us to produce explanations that have more robust recourse, as defined as in \citep{slack2021counterfactual}.

We propose three different variants of the FASTER-CE algorithm to generate explanations reflecting different tradeoffs involving proximity, sparsity, and realism. We theoretically motivate how our method produces more robust explanations. Experiments on three different datasets demonstrate the effectiveness of our approach. We are able to generate multiple explanations with different objectives much faster than existing methods. We also discuss the trade-offs between these objectives experimentally. To the best of our knowledge, this is the first work providing a set of algorithms to generate multiple explanations along key the desirable axes: robustness, sparsity, realism, validity, proximity, and speed.

\section{Related Work}

Counterfactual explanations were originally defined and motivated as a useful means of explaining artificial intelligence models in \citep{wachter2017counterfactual}. Gradient descent in the input space is used in \citep{wachter2017counterfactual} to generate explanations. Subsequently, numerous other methods \citep{mothilal2020explaining,sharma2020certifai,balakrishnan2021towards,barr2021counterfactual,poyiadzi2020face,guidotti2019factual,looveren2021interpretable,joshi2018xgems,keane2021if,pawelczyk2020learning,schleich2021geco} have been proposed to generate these explanations. A recent survey \citep{stepin2021survey} provides details on many such algorithms. However, several challenges exist towards the generation of these explanations \citep{sokol2019counterfactual,barocas2020hidden,artelt2019computation}. For brevity, we only mention  the methods most relevant to our work in this section.

CERTIFAI \citep{sharma2020certifai} provides multiple desirable objectives including proximity, validity, sparsity, and realism of explanations in the generation of counterfactuals. The method uses a genetic algorithm to generate candidate counterfactuals. The use of an evolutionary algorithm allows one to more easily incorporate a variety of objectives or domain constraints compared to gradient based optimization methods, but it can be quite slow, specially for large, high-dimensional data.
DiCE is notable for incorporating a diversity constraint to provide multiple counterfactual explanations, but is also slow, and using gradient descent results in explanations that might not be robust \citep{slack2021counterfactual}.
Latent-CF \citep{balasubramanian2020latent} and Sharpshooter \citep{barr2021counterfactual} are two recent approaches that utilize a latent space (like FASTER-CE) to find counterfactual explanations. Both methods demonstrate the usefulness of representation learning. However, the gradient descent based Latent-CF  is vulnerable to providing explanations that are not robust. Sharpshooter is a fast algorithm, but does not take into account the sparsity or robustness of explanations. Finally, \citet{balakrishnan2021towards} use latent space perturbations to analyze the bias in face recognition algorithms. The method uses such perturbations to generate images for bias analysis, and not towards counterfactual explanations.

\section{Theory}

Given an input $\vec{x}$, a classifier $\vec{f}$, and a distance metric $d$, a counterfactual explanation $\vec{c}$  can be found by solving the optimization problem:
\begin{equation}
\min_{\vec{c}} d(\vec{x},\vec{c}) ~~~
\textrm{s.t.}\; \vec{f}(\vec{x})\neq 
\vec{f}(\vec{c})  
\label{eq:opt}
\end{equation}
However, the solution to such an optimization problem is non-trivial, especially when the classifier is assumed to be a black-box. Moreover, multi-objective extensions for this formulation (where the objectives are beyond proximity and validity) are needed to generate sparse, realistic, and robust explanations. In this section, we describe our method to generate counterfactual explanations. FASTER-CE utilizes the same underlying method and provides three algorithms, each of which has different primary objectives: 1) generating counterfactual explanations with only distance to the input as consideration (where distance is defined based on Euclidean distance for continuous features and simple matching distance for categorical features), 2) generating sparse counterfactual explanations with sparsity as the main objective and distance as a secondary objective, and 3) generating realistic counterfactual explanations with sparsity and distance as secondary objectives. 
A user can choose a specific version of FASTER-CE that best matches her objectives, or simply deploy all three algorithms to generate multiple explanations, since our proposed method is extremely quick. Experimentally, we show that running all three algorithms to generate multiple diverse explanations is still much faster than existing methods that generate just one explanation. 

Consider an autoencoder with an encoder $f_\phi$ and a decoder  $g_\theta$ (Figure \ref{fig:FASTERCE}) that maps input $\textbf{x}$ into a latent space vector $\textbf{z}$ using the encoder. The decoder maps the latent space vector to a reconstructed input $\textbf{x'}$. The autoencoder is trained to minimize the distance (d) based loss = $d(\textbf{x},\; g_{\theta}(f_{\phi}(\textbf{x})))$, 
The distance for continuous features is the normalized Euclidean distance, and for categorical features is one minus the cosine similarity between one-hot encoded vectors.

We first randomly sample the latent space of the trained autoencoder extensively, generate decoder outputs for each of these random samples, and query the black-box model for these decoder outputs. We then calculate the principal axes of variation for the attributes and for the model output. Formally, let there be a list of $n$ attributes (features in the input dataset). We generate a training set $T_{z}$ = $\{z_{i}, x'_{i}\}_{i=1}^{k}$ , where $x'_i$ is the decoder output for latent vector $z_{i}$. The value for attribute $j$ may be continuous i.e. $\in$ [0, 1] or binary i.e. $\in$ {0, 1}.
We produce $T_{z}$ as follows: First, we sample a large number of values of $z_i$
from the latent space of the autoencoder. We obtain the corresponding $x'_{i}$ by passing these perturbations through the decoder. 
For each feature $j$, we use $T_{z}$ to find a $T-1$ dimensional hyperplane $h_{j}=
(n_{j}, b_{j} )$, where $T$ is the dimension of the $T_{z}$, $n_j$ is the normal vector and $b_j$ is the offset. To find these hyperplanes, for continuous features, we train a lasso regression model. For categorical features we train a support vector machine (SVM) classifier with a linear kernel. Simultaneously, we find the prediction of the black-box model $y'$ for each of the decoder outputs corresponding to every randomly generated latent space vector, and train an SVM to find the prediction hyperplane $h_{y'}= (n_{y'}, b_{y'})$. The hyperplane $h_{j}$ (approximately) represents a contour of constant values of feature $j$, and the normal vector $n_j$ specifies the gradient direction for the feature. Similarly, the hyperplane $h_{y'}$ specifies (approximately) a contour of constant values for predictions $y'$, and the normal vector $n_{y'}$ specifies the gradient direction for the prediction.

The feature and prediction hyperplanes are then used to find counterfactual explanations. The method to find counterfactual explanations for a point $x_{test}$ without any consideration for sparsity or user-constraints only depends on the prediction hyperplane and is shown in Algorithm 1. The test point $x_{test}$ is fed into the encoder to generate the latent space vector $z_{test}$. Then, $z_{test}$ is projected onto the hyperplane $h_{y'}$ to get $z_{proj}$. Finally, we use line-search in the direction $n_{y'}$ or $-n_{y'}$ (the direction opposite to the direction of projection) in small incremental steps 
and query the black-box model until a valid counterfactual explanation is found. Linearly increasing increments are used, but multiplicative updates can also be considered. Choosing between $n_{y'}$ and $-n_{y'}$ depends on whether the prediction of the black-box for the test input point matches the prediction of the linear prediction hyperplane. Specifically, if the predictions match, then the decision of the linear hyperplane aligns with the decision of the black-box, and the counterfactual search direction will be in the direction of projection. Otherwise, we move in a direction opposite to the direction of projection until a valid counterfactual explanation is found. Even if the global linear model is unable to approximate well for a highly non-linear black-box model, the counterfactual explanation is likely to remain valid for the original black-box.

\begin{algorithm}[t]

 \caption{FASTER-CE1: Counterfactual Explanation}
  \begin{algorithmic}
   \label{alg:algorithmmoe}
  \STATE Inputs: Test point $x_{test}$. Black-box model $F(x)$, Learned Encoder $f_{\phi}$, Learned Decoder $g_{\theta}$, Learned prediction hyperplane $h_{y'}=(n_{y'},b_{y'})$
  \STATE Hyperparameters: $\epsilon$ the incremental step value , $\delta\epsilon$, $\epsilon_{max}$
  \STATE $z_{test}=f_{\phi}(x_{test})$
  \STATE $z_{proj}$ = Projection of $z_{test}$ on hyperplane $h_{y'}$
  \STATE Initialize $\epsilon$ = 0, $c(\epsilon)$ = $\{\}$ the variable that stores the counterfactual explanations 
  
  \IF{prediction($n_{y'}.f_{\phi}(x_{test}) + b_{y'}) = F(x_{test})$}
    \STATE $s = sign(\frac{z_{proj}}{||z_{proj}||})$ \\ Signed direction vector of projection, direction of counterfactual is same
  \ELSE
    \STATE $s = - sign(\frac{z_{proj}}{||z_{proj}||})$
    \\ If prediction through prediction hyperplane and prediction of input do not match, move in opposite direction to projection
    \ENDIF
  \WHILE{$c(\epsilon)$ is not empty AND $\epsilon$ $\leq$ $\epsilon_{max}$}
    \STATE $z_{proj} = z_{proj} + \epsilon.s.\frac{n_{y'}}{||n_{y'}||}$
    \STATE Append $c(\epsilon)$ with $g_{\theta}(z_{proj})$
    \IF{$F(c(\epsilon)) =  F(x_{test})$}
        \STATE $c(\epsilon)$ = $\{\}$
    \ENDIF
    \STATE $\epsilon$ = $\epsilon$ + $\delta\epsilon$
   \ENDWHILE
\end{algorithmic}

 \end{algorithm}
\subsection{Sparse Counterfactual Explanations}
Algorithm 2 shows the method to produce sparse explanations using the feature hyperplanes along with the prediction hyperplane. For simplicity, we show an example where a sparse explanation is obtained by changing only feature $a$. We project the test point's latent space vector $z_{test}$ onto the intersection between the prediction hyperplane $h_{y'}$ and the feature hyperplane $h_{a}$. To do this, we iteratively project $z_{test}$ onto each of these two hyperplanes till the intersection is found. Then, we find new direction vectors $c_j$ for every feature $j$ such that they are orthogonal to each other (called orthogonalization). Then, to only change feature $a$, we start from the intersection of the hyperplanes $h_{y'}$ and $h_{a}$ and incrementally move in direction $c_a$ or $-c_a$ (dependent on the test points black-box prediction and prediction through the prediction hyperplane), repeatedly querying the black-box model until a valid explanation is found.

Even though we ensure sparsity in the latent space, the decoded output may still have more than one feature change. Such an explanation would be more realistic since the autoencoder has learned the distribution of the dataset. However, if sparsity is an over-riding objective i.e. only one feature is required to be changed, we post-process to try to induce it. Specifically, if more than one feature has changed between the input and counterfactual (i.e. the L0 distance between $c(\epsilon)$ and $x_{test}$ is greater than 1), we set all features except for feature $a$ (the feature required to be sparse) between the input and counterfactual to be equal, and check if this is a valid counterfactual. If it is, this is the final sparse counterfactual, otherwise we perturb $c(\epsilon)$ randomly (in the input space) and check if we can find a valid sparse explanation. This post-processing comes at the cost of speed and potentially the realism of the counterfactual explanation. We discuss this trade-off in the experiments section.

\begin{algorithm}[t]

 \caption{FASTER-CE2: Sparse Counterfactual Explanations}
  \begin{algorithmic}
   \label{alg:algorithmmoe}
  \STATE Inputs: Test point $x_{test}$. Black-box model $F(x)$, Learned Encoder $f_{\phi}$, Learned Decoder $g_{\theta}$, Learned prediction hyperplane $h_{y'}=(n_{y'},b_{y'})$, Learned feature hyperplanes $h_{j}=(n_{j},b_{j})$
  \STATE Hyperparameters: $\epsilon$ the incremental step value , $\delta\epsilon$, $\epsilon_{max}$ 
  \STATE $z_{test}=f_{\phi}(x_{test})$
  \STATE $z_{proj}$ = Projection of $z_{test}$ on intersection of hyperplanes $h_{y'}$ and $h_{a}$ // $a$ is one specific feature
  \STATE Start orthogonalization of feature hyperplane direction vectors: Q,R <- QR Factorization of matrix $[n_{1},n_{2},...,n_{j},...n_{n}]$
  \FOR{$i$ in $\{1,...,n\}$}
    \STATE $c_{i} = n_{i}$ 
    \FOR{$k$ in $\{1,...,n\}$}
        \IF{$i \neq k$}
            \STATE $c_{i} = c_{i} - \frac{Q_{k}.<Q_{k},c_{k}>}{<Q_{k}.Q_{k}>}$
        \ENDIF
    \ENDFOR
  \ENDFOR
  \STATE Initialize $\epsilon$ = 0, $c(\epsilon)$ = $\{\}$ the variable that stores the counterfactual explanations 
  
   \IF{prediction($n_{y'}.f_{\phi}(x_{test}) + b_{y'}) = F(x_{test})$}
    \STATE $s = sign(\frac{z_{proj}}{||z_{proj}||})$ \\ Signed direction vector of projection, direction of counterfactual is same
  \ELSE
    \STATE $s = - sign(\frac{z_{proj}}{||z_{proj}||})$
    \\ If prediction through prediction hyperplane and prediction of input do not match, move in opposite direction to projection
    \ENDIF
  \WHILE{$c(\epsilon)$ is not empty}
    \STATE $z_{proj} = z_{proj} + \epsilon.s.\frac{c_{a}}{||c_{a}||}$
    \STATE Append $c(\epsilon)$ with $g_{\theta}(z_{proj})$
    \IF{$F(c(\epsilon)) =  F(x_{test})$}
        \STATE $c(\epsilon)$ = $\{\}$
    \ENDIF
    \STATE $\epsilon$ = $\epsilon$ + $\delta\epsilon$
   \ENDWHILE
   \STATE Post-processing
   \IF{$||c(\epsilon) - x_{test}||_{0}>1$ }
      \STATE $c(\epsilon) = x_{test}$ for all features except feature $a$
      \IF{$F(c(\epsilon)) = F(x_{test})$}
        \STATE Perturb feature $a$ till $F(c(\epsilon)) \neq F(x_{test})$ 
        \IF{$F(c(\epsilon)) = F(x_{test})$}
            \STATE No valid counterfactual found
        \ELSE
            \STATE Return $c(\epsilon)$
        \ENDIF
    \ENDIF
    \ENDIF
        
\end{algorithmic}

 \end{algorithm}

\subsection{Realistic and Multiple Counterfactual Explanations}

One major concern with Algorithm 2 is that a valid counterfactual that involves changing just one feature, may not exist. Moreover, if highly correlated features are not changed as well, then the explanation may not be realistic. Since the autoencoder's latent space learns the distribution of the data, using perturbations from it can help in producing in-distribution explanations and alleviate this issue.

To provide realistic explanations, user-defined constraints can be incorporated and QR factorization is carried out based on these constraints. For example, if a user cannot substantially change income level  without also changing education status, the user can specify this as a constraint and the FASTER-CE algorithm considers orthogonlization via QR factorization of the normals of the hyperplanes only with respect to features that cannot change. The intersection of hyperplanes is found for features that can change, and the direction vector $c_j$ is only orthogonal to features that cannot change. We show this in Algorithm 3, detailed in the supplementary material. Since FASTER-CE is extremely quick (as shown in the experiments), running the method repeatedly not only with the three proposed algorithms but also with different sets of user constraints using Algorithm 3 can generate multiple explanations with different feature changes and this is still faster than running most other popular methods to generate just one explanation.

\subsection{Robustness of Counterfactual Explanations}

A  notable definition of robustness in the counterfactual explanation literature \citep{dominguez2021adversarial} states that
counterfactual explanations (and hence recourse recommendations) should remain valid (i.e. lead to the same outcome) for all plausible individuals (adequately) similar to the individual seeking recourse.
It has been shown in \citep{dominguez2021adversarial} that for a linear model with a set of actionable features that are unbounded, the existence of robust recourse is guaranteed, and for a linear model, the problem is equivalent to the standard robustness problem for a classifier:
 $   h'(x) = \, <w, x> \, \geq b + ||w^{*}||\epsilon   $ \\
Any recourse recommendation or counterfactual explanation would be adversarially robust by a level of $\epsilon$ i.e. providing more robust explanations is equivalent to having a higher value of $\epsilon$. Since FASTER-CE1 uses a linear hyperplane to find counterfactual explanations and does a line search in the direction of the normal to the prediction hyperplane, we can generate a more robust means of recourse
by making incremental steps in the line-search beyond a counterfactual explanation that may have been optimized based on proximity, sparsity, or realism. Hence, more robust explanations would come at the cost of the distance between the input and a valid counterfactual explanation: the closer the input and explanation are, the closer the explanation is to the decision boundary. In the experiments, we discuss the trade-off between producing more robust explanations and the proximity objective.

\section{Experiments and Results}



Experiments are performed on three different datasets of varying complexity: UCI Adult \citep{kohavi1996scaling}, German Credit \citep{Dua:2019}, and the Lending Club datasets \citep{bworld}. The UCI Adult dataset contains 48842 instances and 14 features, and the model is trained to predict if an individual's income is greater than or less than 50k. The German Credit dataset has 1000 instances and 20 features, and the model is trained to predict an individual's credit risk (good or bad). The Lending Club dataset has 10000 instances and 55 features, and the model is trained to predict loan decisions based on whether an individual will pay back their loan. Each dataset is used to train a multi-layered perceptron neural network which is treated as the black-box model. We consider a neural network as the black-box model since it achieves state of the art accuracies across datasets, and to validate that the proposed approach that approximates these non-linear models linearly can find good counterfactual explanations. We use a standard autoencoder instead of using a variational autoencoder (VAE) since a VAE assumes a gaussian distribution in the latent space, and we did not want any such assumptions for tabular datasets. An autoencoder is trained on each of the three datasets with a 70, 15, 15 split between training set, validation set, and test set. The details of the architectures and performances of the neural networks and autoencoders for each dataset are provided in the supplementary material. This section first reports example counterfactual explanations for each of the FASTER-CE algorithms. Thereafter, we compare our method in terms of various metrics for counterfactual explanations to some state of the art methods and discuss the trade-offs between these metrics. We then show the effect of producing more robust explanations using our algorithm. All experimental results are averaged across five runs and error bars are also provided.


\subsection{Example Counterfactual Explanations}

We show example counterfactual explanations for the UCI Adult dataset in Table 2 for a test point for all three algorithms. We only show the features that changed between the input and counterfactual explanation. FASTER-CE2 is used three times to generate three different types of sparse explanations corresponding to three different features. FASTER-CE3 is used with the constraint that the education and occupation features are the two features that can change. 

The results are shown in Table 2. The test point (input) received a negative decision from the black-box model. When using FASTERCE-1, which only accounts for distance from the prediction hyperplane, the occupation and capital gain features change. When we induce sparsity with respect to a feature using FASTER-CE2, only one feature changes. In the table, we show three examples of just one feature changing. When we include a user-defined constraint that both education and occupation can change, both features change in the resultant counterfactual. For the person subject to the negative decision, any of these decisions could be valuable. Example results for the German Credit dataset and the Lending Club dataset and are provided in the supplementary material. 

\begin{table*}[t!]
\caption{Counterfactual Explanations for the UCI-Adult dataset for all three algorithms. Any feature that has changed with respect to the input is marked in bold}
\centering
\begin{tabular}{cccccc} 
\hline
Features &   Education & Occupation & Capital Gain & Hours-per-week & Prediction \\
\hline
Input    &  Assoc-acdm & Professional & 0 & 40 &  <=50k  \\
FASTER-CE1     &  Assoc-acdm & \textbf{White-collar}  & \textbf{33704} & 40 & >50k \\
FASTER-CE2   & \textbf{Doctorate} &  Professional &   0    & 40 &  >50k   \\
FASTER-CE2   & Assoc-acdm &  Professional &   \textbf{99985}    & 40 &   >50k  \\
FASTER-CE2   & Assoc-acdm &  Professional &   0    & \textbf{76}  &  >50k  \\
FASTER-CE3   & \textbf{Doctorate} &  \textbf{White-collar} &   0    & 40  & >50k   \\

\hline
\end{tabular}
 \label{Table2}

\end{table*}

\subsection{FASTER-CE performance and comparison to other methods}

We compare the three proposed algorithms to three other popular methods that provide counterfactual explanations for all three datasets. All evaluations are carried out on the test sets (randomly chosen) for each of the three datasets. We compare to DiCE \citep{mothilal2020explaining}, CERTIFAI \citep{sharma2020certifai}, Latent-CF \citep{balasubramanian2020latent}, and \citep{looveren2021interpretable} (we call this method ICEGP). While CERTIFAI, DiCE, and ICEGP are three of the most widely used techniques to generate counterfactual explanations, Latent-CF is a recent method that uses gradient descent in the latent space to find counterfactual explanations. Latent-CF requires access to predicted class probabilities and not just the class label, hence is not truly a black-box method.

\begin{figure*}[t!]
  \centering 
  \includegraphics[scale=.3]{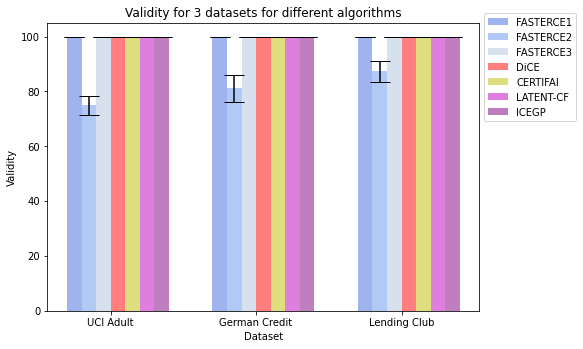}
  \includegraphics[scale=.3]{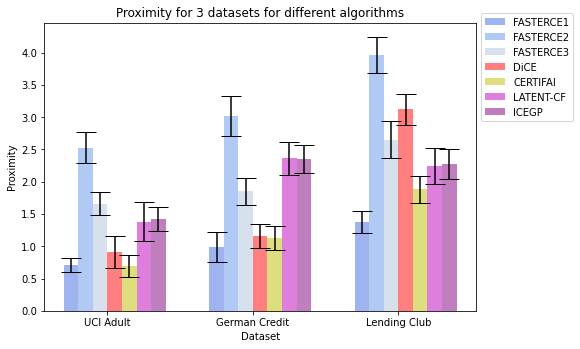} 
  \includegraphics[scale=.3]{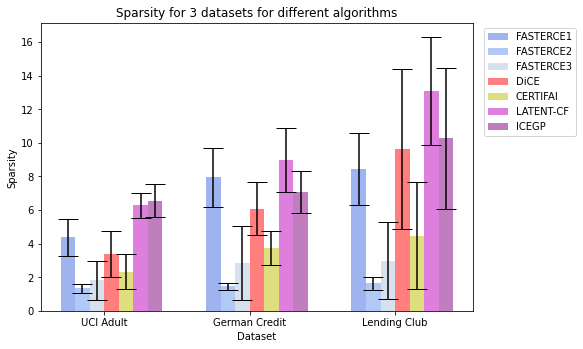}
  \includegraphics[scale=.3]{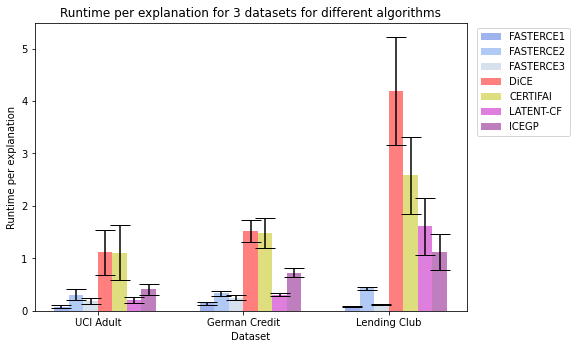}

  \caption{Comparison of methods to FASTER-CE for the UCI Adult dataset, German Credit Dataset, and Lending Club dataset. Validity and sparsity are in percentages. Higher validity is better; sparsity value is lower if less features change and hence a lower value is better; the measure for proximity is just the distance between the input and counterfactual explanation and hence a lower proximity is better, and lower runtime per explanation is better. Time is measured in seconds.}
  \label{fig:german_credit_plots}
\end{figure*} 

We use validity, sparsity, proximity, and efficiency to evaluate the counterfactual explanations found. For validity, we report the percentage of counterfactual explanations that are valid. For sparsity, we compute the average number of features that change between the input and counterfactual explanation for all test points (hence, lower is better). For proximity, the distance between the input and counterfactual explanation is measured by taking the normalized Euclidean distance (normalized by the median absolute deviation) for continuous features and simple matching distance for categorical features, as in \citep{sharma2020certifai}. We report the average distance between the input and counterfactual explanations for all input (test) points. Efficiency is reported via two different types of runtime: total runtime for the algorithm across all test points when the algorithm is run from start to end, and average time per counterfactual explanation generation when the back-bone for any algorithm is already setup. For example, for FASTER-CE, training the autoencoder and training the linear models is a fixed cost with respect to time, and hence in deployment, only the three algorithms as defined in the theory are run. Total runtime results are in the supplementary material. For FASTER-CE2, we average the results that we get by trying to induce sparsity across all features. For FASTER-CE3, we induce the same feature constraints across the test set based on characteristics of the dataset (e.g. gender or race cannot be changed for the UCI Adult dataset). The results for the comparison for all three datasets are shown in Figure 3 for the UCI Adult, the German credit, and the Lending Club datasets.

\begin{table*}[t!]
\caption{Robustness of recourse (in percentage) measured in terms of counterfactual explanations that still have the same prediction after perturbing the input for different values of $\delta\epsilon$ and proximity of counterfactual explanations for the lending club dataset.}
\centering
\begin{tabular}{ccc} 
\hline
Metric &   Robustness & Proximity \\
\hline
FASTER-CE1 $\delta\epsilon=0.05$  &86.59 $\pm$ 2.54  & 1.29 $\pm$ 0.22    \\
FASTER-CE1 $\delta\epsilon=0.1$     & 88.70 $\pm$ 3.19 &  1.38 $\pm$ 0.18   \\
FASTER-CE1 $\delta\epsilon=0.3$   & \textbf{100 $\pm$ 0} & 4.08 $\pm$ 0.34      \\
FASTER-CE1 $\delta\epsilon=0.5$   & \textbf{100 $\pm$ 0} & 6.41 $\pm$ 0.28      \\
FASTER-CE1 $\delta\epsilon=1$   & \textbf{100 $\pm$ 0} & 13.43 $\pm$ 0.67     \\
DiCE   & 93.41 $\pm$ 2.55  &  3.12 $\pm$ 0.27  \\
CERTIFAI   & 89.72 $\pm$ 2.93&  1.88 $\pm$ 0.36  \\
Latent-CF   & 90.51 $\pm$ 3.21&  2.24 $\pm$ 0.20 \\

\hline

\end{tabular}
 \label{Table4}
\end{table*}

The validity of explanations is  100\% for all methods except FASTER-CE2. This is because the severe restriction of changing just one feature may not yield a valid counterfactual explanation. 
This is also why the time taken for FASTER-CE2 is higher: trying to find a valid explanation takes longer. However, explanations produced by FASTER-CE2 are more sparse compared to any of the other methods across all three datasets. 

CERTIFAI performs well in terms of proximity for datasets with lesser number of features (UCI Adult and German Credit) and FASTER-CE1 performs comparably. However, with a larger dimensional dataset (Lending Club), FASTER-CE1 performs better. This is because CERTIFAI uses a genetic algorithm that does not scale well as the dimensionality of the dataset grows when the number of generations for the algorithm is restricted. Our method relies on projections from the lower dimensional latent space, and hence is able to find explanations that are closer much faster. 

Across all datasets, FASTER-CE is much faster for runtime per explanation, which is crucial for explainability methods in deployment. Across datasets, FASTER-CE2 and FASTER-CE3 are able to find the intersection between hyperplanes with a maximum of 20 iterations. FASTER-CE3 is able to provide a reasonable trade-off between sparsity, validity, and the time taken to generate explanations. The choice of which algorithm to use is up to the end-user based on explanation needs. Since the time taken for generating an individual explanation across all FASTER-CE algorithms is still significantly less than producing one explanation for other methods for a larger dataset, a user can run all algorithms and choose the means of recourse that is best for them.


{\bf Robust Explanations}. To examine the robustness of FASTER-CE based explanations 
we randomly perturb the features in the input that changed between the input and counterfactual explanation found using FASTER-CE1, and find the corresponding counterfactual explanation using FASTER-CE1. We report the proximity and robustness of recourse (measured by checking if the recourse remains valid) for the counterfactual explanation 
(a robust recourse would keep the counterfactual explanation valid after perturbation). We do this for different values of $\delta\epsilon$, and report the results for the lending club dataset, and compare to DiCE, CERTIFAI and Latent-CF.
The results are in Table 5. FASTER-CE is able to produce explanations that are more robust to input manipulations. This comes at the cost of proximity: producing more robust explanations results in the input and counterfactual explanation being further away. However, the proximity of explanations is  not much  compromised and in practice one may prefer the robust explanation version for this reason.

\section{Conclusion and Future Work}

This paper introduces FASTER-CE, a novel set of algorithms to generate counterfactual explanations for a black-box model
based on multiple quality objectives. 
This work was motivated by actual business applications that demanded much faster 
 generation of explanations that were likely actionable based on multiple criteria - a need that was not met by any existing algorithm.
FASTER-CE can produce sparse, realistic, and robust explanations efficiently. 
For future work, we would like to examine the tradeoff in
building multiple (locally) linear models for different points as opposed to training one global linear model per target variable. This will take more time, but may produce even better counterfactuals.
Further, we would be interested in studying the effect of adding a realism constraint beyond providing user-defined constraints, such as including an objective that tries to optimize the distance between a counterfactual explanation and class prototypes in the z-space. It would also be interesting to investigate the variability in explanations with drift in data. While the total runtime for FASTER-CE is still less than existing methods (which also do not consider drift) and hence it would still be faster for new data, we would like to extend the training procedure to account for such streaming data.

\bibliographystyle{ACM-Reference-Format}
\bibliography{main.bib}

\newpage
\appendix
\section{Supplementary Material}
Additional experiments and results, along with a discussion on the method is provided in this section.

\subsection{Realistic and Multiple Counterfactual Explanations}

One major concern with Algorithm 2 is that a valid counterfactual that involves changing just one feature, may not exist. Moreover, if highly correlated features are not changed as well, then the explanation may not be realistic. Since the autoencoder's latent space learns the distribution of the data, using perturbations from it can help in producing in-distribution explanations and alleviate this issue.

To provide realistic explanations, user-defined constraints can be incorporated and QR factorization is carried out based on these constraints. For example, if a user cannot substantially change income level  without also changing education status, the user can specify this as a constraint and the FASTER-CE algorithm considers orthogonlization via QR factorization of the normals of the hyperplanes only with respect to features that cannot change. The intersection of hyperplanes is found for features that can change, and the direction vector $c_j$ is only orthogonal to features that cannot change. We show this in Algorithm 3 for a case when the constraint is based on two features, but the method can easily be extended for any such constraints. Moreover, if the values associated with any feature need to be within a range, the corresponding latent space vectors can be generated by random sampling such that latent vectors that have decoder outputs beyond this range are not considered to train the linear model. If user-defined constraints are not available, this algorithm can be run repeatedly for different combinations of feature changes to generate multiple diverse counterfactual explanations. Since FASTER-CE is extremely quick, running the method repeatedly is still faster than running other existing methods to generate one explanation (as shown in the experiments). 

\begin{algorithm}[t!]

 \caption{FASTER-CE3: Realistic Counterfactual Explanations}
  \begin{algorithmic}
   \label{alg:algorithmmoe}
  \STATE Inputs: Test point $x_{test}$. Black-box model $F(x)$, Learned Encoder $f_{\phi}$, Learned Decoder $g_{\theta}$, Learned prediction hyperplane $h_{y'}=(n_{y'},b_{y'})$, Learned feature hyperplanes $h_{j}=(n_{j},b_{j})$, User-defined constraints: Only change features $1$ and $2$
  \STATE Follow Algorithm 2 till projection step
  \STATE Start orthogonalization of feature hyperplane direction vectors: Q,R <- QR Factorization of matrix $[n_{1},n_{2},...,n_{j},...n_{n}]$
  \FOR{$i$ in $\{1,...,n\}$}
    \STATE $c_{i} = n_{i}$ 
    \\ Adding user-defined constraints
    \FOR{$k$ in $\{1,...,n\}$}
        \IF{($i =1$ AND $k=2$) OR ($i=2$ AND $k=1$)}  
         \STATE $c_{i} = n_{i}$
        \ELSIF{$i \neq k$}
            \STATE $c_{i} = c_{i} - \frac{Q_{k}.<Q_{k},c_{k}>}{<Q_{k}.Q_{k}>}$
        \ENDIF
    \ENDFOR
  \ENDFOR
  \STATE Initialize $\epsilon$ = 0, $c(\epsilon)$ = $\{\}$ the variable that stores the counterfactual explanations 
  
   \IF{prediction($n_{y'}.f_{\phi}(x_{test}) + b_{y'}) = F(x_{test})$}
    \STATE $s = sign(\frac{z_{proj}}{||z_{proj}||})$ \\ Signed direction vector of projection, direction of counterfactual is same
  \ELSE
    \STATE $s = - sign(\frac{z_{proj}}{||z_{proj}||})$
    \\ If prediction through prediction hyperplane and prediction of input do not match, move in opposite direction to projection
    \ENDIF
  \WHILE{$c(\epsilon)$ is not empty}
    \STATE $z_{proj} = z_{proj} + \epsilon.s.\frac{c_{j}}{||c_{j}||}$
    \STATE Append $c(\epsilon)$ with $g_{\theta}(z_{proj})$
    \IF{$F(c(\epsilon)) =  F(x_{test})$}
        \STATE $c(\epsilon)$ = $\{\}$
    \ENDIF
    \STATE $\epsilon$ = $\epsilon$ + $\delta\epsilon$
   \ENDWHILE
\end{algorithmic}
\label{algorithmmoe}

 \end{algorithm}

\subsection{Autoencoder and Black-Box details}

The details of the architectures and the accuracies of the neural networks and autoencoders for each dataset are shown in Table 1. For every dataset, the categorical features are one-hot encoded and the continuous features are normalized. The architectures for both the neural networks and the autoencoders for each dataset have been experimented upon and the values associated with the best accuracy and best loss (lower is better), respectively, are reported.

\subsection{Example Counterfactual Explanations}

Results similar to the UCI Adult dataset are observed for the German Credit dataset and the Lending Club dataset, as shown in Table \ref{Tablegerman} and Table \ref{Tablelending} respectively. Each of these explanations are valid counterfactuals, and can be used as a means of recourse. Moreover, each of the explanations is close to the input, and has different characteristics.

\subsection{Total runtime}

The total runtime (Figure \ref{fig:german_credit_plots} across methods is comparable for lower dimensional datasets. However, the total time for the higher dimensional dataset increases drastically for other methods, as does the individual average time as shown in the main paper. 

\begin{table*}[t!]
\caption{Counterfactual Explanations for the German Credit dataset for all three algorithms. Any feature that has changed with respect to the input is marked in bold. For ease of understanding, categorical feature values have been changed in the . For status of checking account: Low - A11, Medium - A12, and High - A13 in the dataset. For payment status of previous credit: All credits paid back duly - A31, Delay - A33}
\centering
\begin{tabular}{cccccc} 
\hline
Features &   Status & Duration & Credit amount & Previous credit & Prediction \\
\hline
Input    &  Low & 36 & 6887 & Delay &  Bad  \\
FASTER-CE1     &  \textbf{Medium} & \textbf{24}  & 6887 & \textbf{Paid back duly} & Good \\
FASTER-CE2   & \textbf{High} &  36 &   6887    & Delay &  Good   \\
FASTER-CE2   & Low &  36 &   6887    & \textbf{Paid back duly} &   Good  \\
FASTER-CE3   & \textbf{Medium} &  \textbf{16} &   6887    & Delay  & Good   \\

\hline
\end{tabular}
 \label{Tablegerman}

\end{table*}

\begin{table*}[t!]
\caption{Counterfactual Explanations for the Lending Club dataset for all three algorithms. Any feature that has changed with respect to the input is marked in bold}
\centering
\begin{tabular}{ccccc} 
\hline
Features &   Interest Rate & FICO & Debt-To-Income Ratio & Prediction \\
\hline
Input    &  0.1071 & 707 & 14.29 &  Fully Paid  \\
FASTER-CE1     &  0.1071 & \textbf{682}  & \textbf{15.13} &  Charged Off \\
FASTER-CE2   & \textbf{0.193} &  707 &  14.29    &  Charged Off   \\
FASTER-CE2   & 0.1071 &  \textbf{614} &   14.29    &   Charged Off  \\
FASTER-CE3   & 0.1071 & \textbf{682} &   \textbf{15.13}  & Charged Off   \\

\hline
\end{tabular}
 \label{Tablelending}

\end{table*}

\begin{figure*}[t!]
  \centering
  
    \includegraphics[scale=.5]{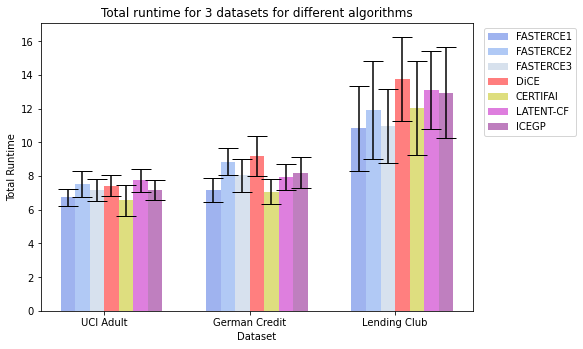}
  \caption{Comparison of methods to FASTER-CE for the UCI Adult dataset, German Credit Dataset, and Lending Club dataset. Time is measured in seconds.}
  \label{fig:german_credit_plots}
\end{figure*} 

\subsection{Discussion}

As shown in the experiments above, the three proposed algorithms can generate a diverse set of explanations efficiently and these explanations are valid, close to the input, sparse, realistic, and robust. However, there are trade-offs among all these objectives, as demonstrated by the results in Figure 3 and Table 5. $\delta\epsilon$ controls the trade-off between not only robustness and proximity, but also runtime. For the results in Figure 3, we use $\delta\epsilon=0.1$ across datasets, and find that a valid counterfactual explanation is returned in not more than 20 iterations in most cases for FASTER-CE1 and FASTER-CE3. A larger value of $\delta\epsilon$ might lead to finding a valid explanation quickly that is also more robust, compared to using a smaller value, but the counterfactual explanation will be further away from the input. Furthermore, as shown by FASTER-CE2, sparsity as a concrete objective can lead to producing invalid counterfactuals. The runtime of the algorithm to produce sparse explanations is also more than the other FASTER-CE algorithms because of the post-processing step. Experimentally, we have found that a dataset with more categorical features that have fewer categories (eg. a binary protected attribute for gender) lead to more invalid explanations. Moreover, the proximity of these explanations on average are also worse than when the sparsity constraint is not considered. While we generate individually realistic explanations by including user constraints, evaluating the realism of an explanation quantitatively can be done at a global level by comparing the distribution of the counterfactual explanations to the input dataset. We refrain from doing this evaluation because a counterfactual explanation is local in nature, and how realistic it is should be determined by an end-user subject to the decision of the model. 

We use the definition of robustness of recourse from \citep{dominguez2021adversarial}, which defines robustness in terms of the validity of recourse. Given this definition, we show that FASTER-CE algorithms can easily provide more robust recourse than existing methods, and to the best of our knowledge, FASTER-CE is the only existing counterfactual explanation method that tries to provide robust recourse with respect to the definition in \citep{dominguez2021adversarial}. However, there can be other possible notions of robustness. For example, similar individuals should achieve similar recourse not only in terms of validity, but also proximity, sparsity, and realism. Investigating robustness of recourse in terms of all these factors is an important and open problem. 

It is also interesting to note that this method may potentially lend itself towards understanding the bias of a black-box model. Specifically, if the linear hyperplane associated with a protected attribute like gender has a normal vector that is perpendicular to the normal vector of the prediction hyperplane, this means that the prediction is not impacted directly by the protected attribute. Furthermore, if the normal vector of the protected attribute hyperplane is orthogonal to all other features, this protected attribute also doesn't effect the values of other features in the dataset. However, given that we are training an autoencoder to learn the distribution of data, and then approximating the black-box with a linear hyperplane, such an analysis needs to be approached with caution. Even though these approximations don't significantly effect the generation of a counterfactual explanation owing to the conditional constraints to find the direction of constant search and constant querying of the black-box model in our algorithms, the fairness of a model might still differ between a linear approximation of the model and the original black-box.


\end{document}